# DeLLiriuM: A large language model for delirium prediction in the ICU using structured EHR

Miguel Contreras[1,6], Sumit Kapoor[2], Jiaqing Zhang[3,6], Andrea Davidson[4,6], Yuanfang Ren[4,6], Ziyuan Guan[4,6], Tezcan Ozrazgat-Baslanti[4,6], Subhash Nerella[1,6], Azra Bihorac[4,6], Parisa Rashidi[1,6]*

[1]Department of Biomedical Engineering, University of Florida, Gainesville, FL, USA
[2]Department of Critical Care Medicine, University of Pittsburgh, Pittsburgh, PA, USA
[3]Department of Electrical and Computer Engineering, University of Florida, Gainesville, FL, USA
[4]Division of Nephrology, Department of Medicine, University of Florida, Gainesville, FL, USA
[5]Department of Surgery, University of Florida, Gainesville, FL, USA
[6]Intelligent Clinical Care Center (IC3), University of Florida, Gainesville, FL, USA

**\* Correspondence:**

**Parisa Rashidi**
**parisa.rashidi@ufl.edu**

**Abstract**

Delirium is an acute confusional state that has been shown to affect up to 31% of patients in the intensive care unit (ICU). Early detection of this condition could lead to more timely interventions and improved health outcomes. While artificial intelligence (AI) models have shown great potential for ICU delirium prediction using structured electronic health records (EHR), most of them have not explored the use of state-of-the-art AI models, have been limited to single hospitals, or have been developed and validated on small cohorts. The use of large language models (LLM), models with hundreds of millions to billions of parameters, with structured EHR data could potentially lead to improved predictive performance. In this study, we propose DeLLiriuM, a novel LLM-based delirium prediction model using EHR data available in the first 24 hours of ICU admission to predict the probability of a patient developing delirium during the rest of their ICU admission. We develop and validate DeLLiriuM on ICU admissions from 104,303 patients pertaining to 195 hospitals across three large databases: the eICU Collaborative Research Database, the Medical Information Mart for Intensive Care (MIMIC)-IV, and the University of Florida Health's Integrated Data Repository. The performance measured by the area under the receiver operating characteristic curve (AUROC) showed that DeLLiriuM outperformed all baselines in two external validation sets, with 0.77 (95% confidence interval 0.76-0.78) and 0.84 (95% confidence interval 0.83-0.85) across 77,543 patients spanning 194 hospitals. To the best of our knowledge, DeLLiriuM is the first LLM-based delirium prediction tool for the ICU based on structured EHR data, outperforming deep learning baselines which employ structured features and can provide helpful information to clinicians for timely interventions.

# 1. Introduction

Delirium is an acute confusional state characterized by fluctuating course, attention deficits, and severe disorganization of behavior [1] that has been shown to affect up to 31% of patients in the intensive care unit (ICU) [2]. Delirium is shown to be associated with longer ICU and hospital stays, as well as higher ICU and in-hospital mortality rates [3]. Current methods for delirium diagnosis are limited to manual assessments such as the Confusion Assessment Method for the ICU (CAM-ICU) and the Intensive Care Delirium Screening Checklist (ICDSC) [4]. Although these methods have shown high diagnostic accuracy in the critical care setting [5], they can only detect delirium once the patient has developed it. Early detection of this condition could lead to more timely interventions and improved health outcomes.

Multiple studies have developed and validated early detection tools for delirium. The PRE-DELIRIC and E-PRE-DELIRIC models, both based on multivariate logistic regression models, were developed and externally validated for delirium prediction using risk factors available in the first 24 hours of a patient's ICU admission [6], [7]. Other studies have focused on using derived predictive features from electronic health records (EHR) to predict delirium at any point of ICU admission using machine learning (ML) classification models [8]. Dynamic delirium prediction models have also been developed to provide continuous risk prediction with 12-24 hours of anticipation using EHR temporal data (such as vital signs, laboratory test results, assessment scores, and medications) from the previous 12-24 hours along with static data (such as age, gender, comorbidities) obtained at admission [9], [10]. However, these models are limited due to relatively small study cohorts in the development of the PRE-DELIRIC and E-PRE-DELIRIC models (around 3,000 patients in each) [6], [7]. Consequently, the generalizability of the models to larger populations is constrained. On the other hand, studies with larger cohorts are limited to validating results on single centers [8], [10] or use classification ML and deep learning models (such as Gated Recurrent Unit [11], Categorical Boosting [12], Recurrent Neural Networks [13]) which are limited in capturing the long context of EHR data [8], [9], [10]. This limitation can be overcome by using state-of-the-art (SOTA) artificial intelligence (AI) models which are proven to capture long-range dependencies [14].

Large language models (LLM) have gained great interest in the healthcare field [15]. These models have a massive number of parameters, ranging from hundreds of millions to billions, and have shown impactful results in tasks requiring human language interpretation. With the objective of improving performance in medical/clinical tasks, different LLMs have been specifically developed for these domains [16], [17], [18]. Particularly, the potential of using such models for clinical outcome predictions has been explored for multiple tasks such as in-hospital mortality [19], heart failure [20], and ICU length of stay [21]. However, most of these studies have employed clinical notes written by healthcare professionals. Few studies have explored the potential of also integrating EHR structured data in text form, achieving improved performance compared to structured features deep learning approaches [20], [22], [23]. Therefore, using LLMs with EHR structured data for delirium prediction has the potential to improve predictive performance.

In this study we propose DeLLiriuM, a novel LLM-based delirium prediction tool which employs structured EHR data in text form. We develop and validate DeLLiriuM on ICU admissions from 104,303 patients pertaining to 195 hospitals across three large databases: the eICU Collaborative Research Database [24] (hereafter referred to as eICU), the Medical Information Mart for Intensive Care (MIMIC)-

IV [25] (hereafter referred to as MIMIC), and the University of Florida Health's Integrated Data Repository (hereafter referred to as UFH). To the best of our knowledge, DeLLiriuM is the first LLM-based delirium prediction tool for the ICU based on structured EHR data.

The study's main contributions are summarized as follows:

1. We propose DeLLiriuM, a novel LLM-based delirium prediction model using structured EHR data available in the first 24 hours of ICU admission to predict the probability of a patient developing delirium during the rest of the ICU admission.
2. We design a pipeline for converting structured EHR into a text report format compatible with LLM models.
3. We propose a novel approach for interpretability of text classification outputs compatible with LLM models.

## 2. Methods

2.1 Data and Study Design

Three databases were used in this study: UFH, MIMIC, and eICU (cohort diagrams in Fig. 1). All data were collected retrospectively. The UFH dataset was retrieved from the University of Florida (UF) Integrated Data Repository and included adult patients admitted to the ICUs at the UF Health Shands Hospital Gainesville location between 2014 to 2019. The MIMIC dataset is a publicly available dataset collected at the Beth Israel Deaconess Medical Center from 2008 to 2019 [25]. The eICU dataset contains data from ICU patients in 208 hospitals in the Midwest, Northeast, South, and West regions of the US from 2014 to 2015 [24]. In all three datasets, patients less than 18 years of age were excluded, and ICU admissions were excluded if they were not the first ICU admission of the patient recorded in the dataset and/or had a length of stay less than 24 hours. These criteria were used to avoid a potential bias towards predicting higher delirium risk during subsequent ICU admissions and to provide enough data for predictions. Similarly, to avoid bias towards predicting higher delirium risk in patients with higher acuity on admission, patients who passed away within 48 hours of admission, and/or presented delirium or coma in the first 24 hours of ICU admission were also excluded. Finally, ICU admissions were excluded if no EHR data was present for the first 24 hours. The UFH dataset was used for training, tuning, and as internal validation sets. The MIMIC and eICU datasets were used as external validation sets to evaluate the generalization of the model to diverse hospital settings. The cohort selection process is shown in Fig. 1.

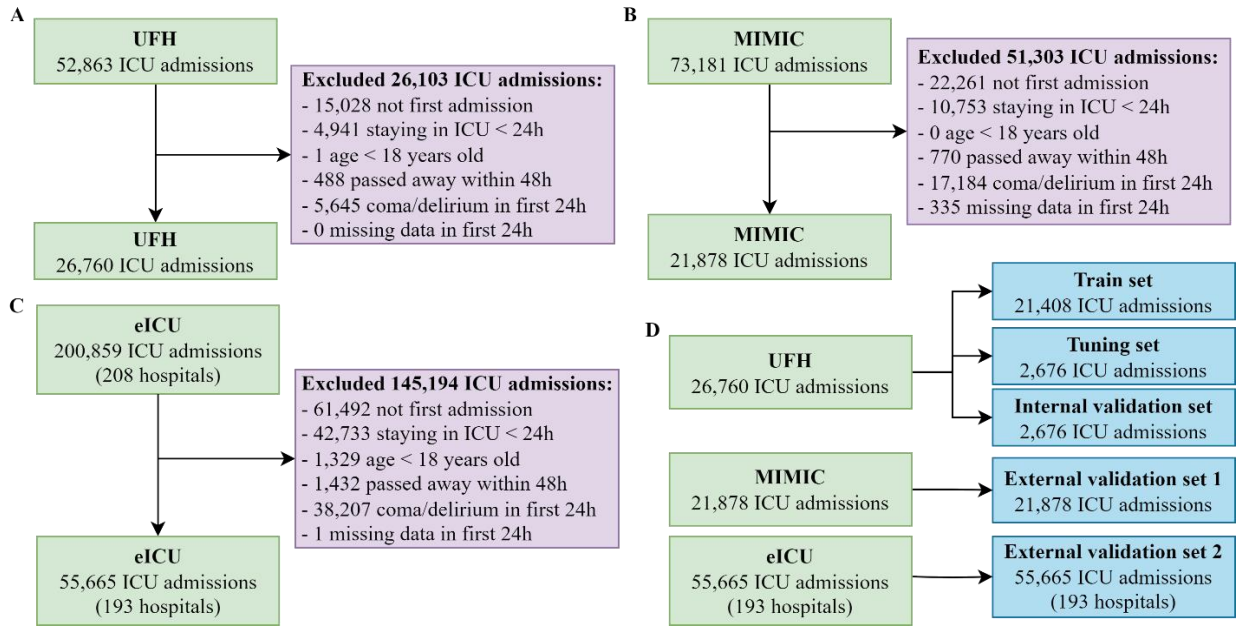

**Fig. 1 | Cohort flow diagram.** The (A) eICU, (B) MIMIC, and (C) UFH datasets. (D) Final datasets were assembled from the three datasets for training, tuning hyperparameters, internal validation, and two external validations. We created the training, tuning, and internal validation sets by splitting the UFH dataset. The two external validation sets were created from the MIMIC and eICU datasets.

2.2 Ethics Approval and Patient Consent

Data from UFH were obtained with approval from the University of Florida Institutional Review Board (IRB) as an exempt study. Subjects were enrolled via waiver of informed consent (IRB201901123). The analysis using the eICU dataset is exempt from IRB board approval due to the retrospective design, absence of direct patient intervention, de-identification of data, and security schema. The data in the MIMIC dataset is de-identified, and the IRBs of the Massachusetts Institute of Technology and Beth Israel Deaconess Medical Center both manage the data repository and have approved the database for external research purposes.

2.3 Outcomes and Features

The primary outcome predicted by our DeLLiriuM model is the risk of developing delirium at any point during a patient's ICU admission after 24 hours. The presence of delirium is defined as a positive CAM score along with a Richmond Agitation Sedation Scale (RASS) score of -3 or higher [26] at any 12-hour interval after 24 hours of ICU admission (Fig. 2).

To predict the outcome, ICU temporal data and static patient information were used as predictive features. Temporal data was extracted from the first 24 hours of ICU admission and was composed of four categories of variables: vital signs, laboratory test measurements, medications, and assessment scores (Fig. 2). Static data comprised demographic and comorbidity information and was extracted from patient admission information (Fig. 2). A total of 81 predictive features, common in all study cohorts, were used for delirium prediction. A complete list of the variables used for prediction can be found in the Appendix (Table A1).

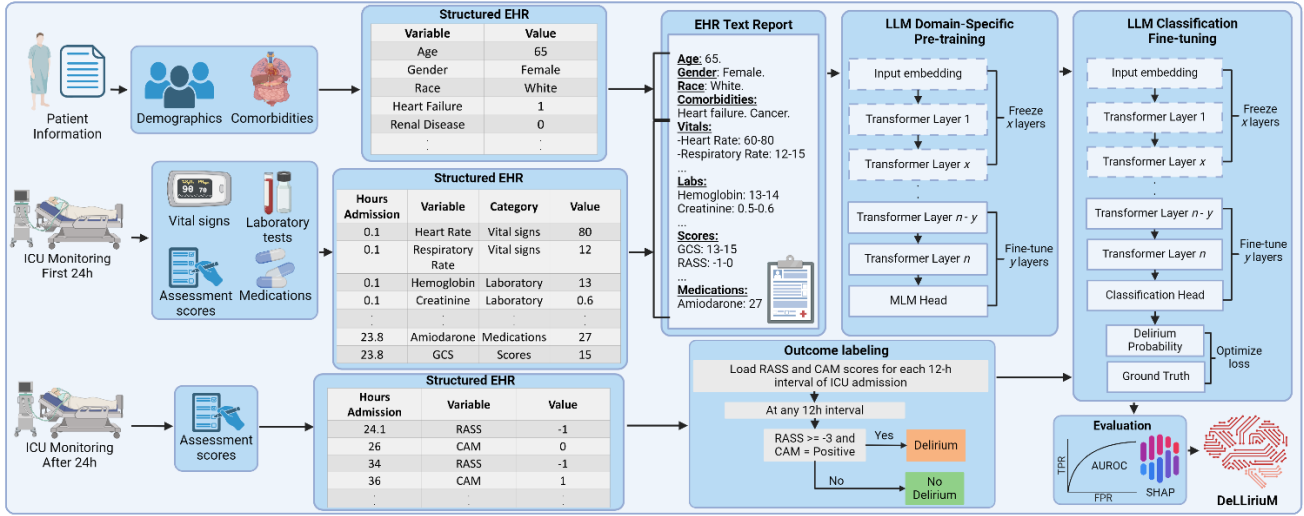

**Fig. 2 | DeLLiriuM model development overview.** Data from three different Intensive Care Unit (ICU) datasets were used for training and validating the DeLLiriuM model (195 hospitals, 104,303 patients). Each dataset consisted of ICU monitoring data (*i.e.*, vital signs, medications, laboratory test results, and assessment scores) and patient admission information (*i.e.*, patient demographics and comorbidities). Two assessment scores were extracted for assessment of delirium after the first 24 hours of ICU admission: RASS and CAM. Delirium was defined as any 12-hour interval in which the lowest RASS score was greater or equal to -3 along with at least one positive CAM score. Temporal data from the first 24 hours of ICU monitoring was summarized by taking the minimum and maximum values *(i.e.,* range) of each variable and converted to text along with static patient admission data. GatorTronS [27] was used as the backbone for the model, which was first pre-trained on the generated EHR text reports with a masked language modeling (MLM) objective and then fine-tuned with a classification objective for delirium prediction. Finally, the model was evaluated using Area Under the Receiver Operating Characteristic (AUROC) and SHapley Additive exPlanations (SHAP) analysis for interpretability, yielding the final DeLLiriuM model.

## 2.4 Model Development and Performance

The DeLLiriuM model was trained using 80% of the UFH dataset. The remaining 20% was used for tuning hyperparameters (10%) and for internal validation (10%). The data split was performed randomly while ensuring no patient data was common between the three data partitions. The data processing for DeLLiriuM involved generating a text report from the static and temporal data variables. The DeLLiriuM model used the GatorTronS [27], a clinical 345 million-parameter LLM, as its backbone with a context length of 512. This model was chosen as the backbone given it yielded the best performance across all cohorts compared to other LLMs (including GatorTron-8b [18], Meditron-7b [16], and LLaMa 3-8b [28]). To adhere to this sequence length limit, the temporal data was summarized by taking the minimum and maximum values of each variable. In the case of medications, the total dose given in the first 24 hours was used. The static data was then appended to the beginning of the report (Fig. 2).

For the training process, domain-specific pre-training was first conducted by using the summarized EHR text reports from the train set with a Masked Language Modelling (MLM) objective [29]. The pre-training was conducted by keeping the first *x* number of layers frozen and training the last *y* number of layers along with the MLM head (Fig. 2). Both *x* and *y* were used as optimizable hyperparameters along with learning rate and batch size. To determine the optimal hyperparameters, the Optuna library [30] was used to run 20 training trials. Each trial was set to 100 epochs and the best model (*i.e.,* epoch with best evaluation metric) was loaded at the end. The trial with the lowest loss on the tuning set resulted in the domain-specific pre-trained model. Then, the domain-specific pre-trained model was fine-tuned for the delirium classification task by keeping the first *x* number of layers frozen and training the last *y* number of layers along with the classification head (Fig. 2). Similar to the pre-training, both *x* and *y* were used as optimizable hyperparameters along with learning rate and batch size, and the Optuna library was used to run 20 training trials. Each trial was set to 30 epochs and the best model (*i.e.,* epoch with best evaluation metric) was loaded at the end. The trial with the best Area Under the Receiver Operating Characteristic (AUROC) on the tuning set resulted in the final DeLLiriuM model. The parameter search space and final parameters for pre-training and fine-tuning can be found in the Appendix (Table A2).

Two types of baselines were employed for comparison with DeLLiriuM: structured EHR and text EHR. The structured EHR baseline consisted of three deep learning models: Neural Network (NN), Transformer [31], and Mamba [32]. The NN model employs five statistical features (*i.e.,* mean, standard deviation, minimum, maximum, and missingness indicator) to convert the temporal data into a static representation and concatenate with the static data. The Transformer and Mamba models are based on architectures previously developed for clinical outcome predictions [14], [33], [34]. These models convert the time series data into embeddings, with each variable represented as a token, which is then fused with the static data to predict the outcome. The text EHR baseline consisted of four LLM models: ClinicalBERT [17], GatorTron-8b [18], GatorTronS [27], and LLaMa 3-8b [28]. Each LLM was fine-tuned only on classification following the same strategy as DeLLiriuM.

2.5 Model Interpretability

For interpretation of the DeLLiriuM model predictions, SHapley Additive exPlanations (SHAP) analysis [35] was employed. Given that the input to the model is text, summarizing the importance of feature numeric values can be challenging. To address this challenge, we introduced a novel approach to conduct SHAP analysis on text input for classification. First, the SHAP analysis was conducted in the entire EHR text reports. Then, each report was sectioned by feature using the '[SEP]' token and a label for each feature was generated. The sum of the absolute SHAP values for each section was computed, and then, the mean SHAP value across all samples for each feature was taken. Finally, the mean SHAP value associated with each feature was used to generate a bar plot. The approach is summarized in Fig. 3. Furthermore, four random examples were used to visualize the SHAP text plot of positive and negative delirium examples with low and high DeLLiriuM scores.

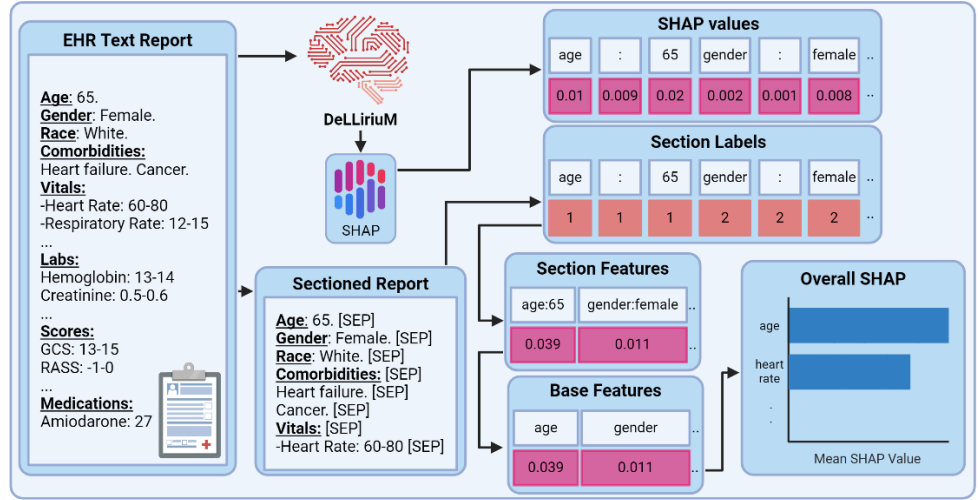

**Fig. 3 | SHAP analysis approach for DeLLiriuM model.** The model predictions were interpreted using SHAP analysis. To obtain the overall SHAP value of each feature, text reports were sectioned using the '[SEP]' token and a label was generated for each section. The sum of the absolute SHAP values for each section was computed, and the mean SHAP value across all samples for each feature was taken.

2.5 Statistical Analysis

To determine if the difference in performance between baseline models and the DeLLiriuM model was statistically significant, the AUROC values between algorithms were compared using a Wilcoxon rank sum test. A 200-iteration bootstrap was performed to calculate the 95% confidence interval (CI), and the median across the bootstrap was used to represent the overall AUROC value.

## 3. Results
### 3.1 Patient Characteristics

Patient characteristics for this study are presented in Table I for all cohorts in terms of ICU admissions. Given that only the first ICU admission of each patient was considered for this study, the number of ICU admissions is equal to the number of patients. Delirium incidence, as defined by the diagnostic criteria of this study, was 3.6% in UFH (982 patients), 6.8% in MIMIC (1,502 patients), and 2.1% in eICU (1,189 patients). Overall, delirium incidence across all cohorts was 3.5% (3,673 patients).

Across all cohorts, patients with delirium had a higher median age, lower body mass index (BMI), longer ICU admissions, and higher coma and mortality rates compared to patients without delirium. In terms of comorbidities in delirious patients, the UFH cohort showed higher chronic heart failure (CHF), renal disease and liver disease rates. The MIMIC cohort showed higher chronic obstructive pulmonary disease (COPD), cerebrovascular accident (CVA), and liver disease rates. The eICU cohort showed higher CVA and human immunodeficiency virus (HIV) rates.

**Table 1. Patient characteristics for three study cohorts.**

| Cohort | UFH | | MIMIC | | eICU | |
|---|---|---|---|---|---|---|
| Item | Non-Delirium (*n* = 25,778) | Delirium (*n* = 982, 3.6%) | Non-Delirium (*n* = 20,376) | Delirium (*n* = 1,502, 6.8%) | Non-Delirium (*n* = 54,476) | Delirium (*n* = 1,189, 2.1%) |
| **Basic information** | | | | | | |
| Hospitals, n | 1 | 1 | 1 | 1 | 193 | 45 |
| Age, years, median (IQR) | 62.0 (49.0-72.0) | 67.0 (57.0-76.0)* | 65.0 (52.0-77.0) | 70.0 (59.0-81.0)* | 66.0 (54.0-77.0) | 70.0 (58.0-81.0)* |
| Female, n (%) | 11,845 (46.0%) | 391 (39.8%)* | 9,613 (47.2%) | 693 (46.1%) | 25,404 (46.6%) | 558 (46.9%) |
| BMI, kg/m2, median (IQR) | 27.4 (23.4-32.4) | 26.6 (23.2-31.4)* | 27.5 (23.9-32.2) | 27.1 (23.3-31.6)* | 27.7 (23.7-33.0) | 26.8 (22.8-31.6)* |
| ICU length of stay, days, median (IQR) | 3.0 (1.9-5.1) | 9.1 (5.2-15.0)* | 2.0 (1.4-3.1) | 5.9 (3.5-9.8)* | 2.1 (1.5-3.5) | 4.9 (2.9-8.6)* |
| CCI, median (IQR) | 2.0 (0.0-3.0) | 2.0 (1.0-4.0)* | 3.0 (1.0-5.0) | 3.0 (2.0-5.0)* | 1.0 (1.0-2.0) | 1.0 (1.0-2.0) |
| **Race, n (%)** | | | | | | |
| Black | 4,246 (16.5%) | 129 (13.1%)* | 2,000 (9.8%) | 138 (9.2%) | 6,398 (11.7%) | 187 (15.7%)* |
| White | 19,806 (76.8%) | 794 (80.9%)* | 14,321 (70.3%) | 1,017 (67.7%) | 42,240 (77.5%) | 886 (74.5%)* |
| Other | 1,726 (6.7%) | 59 (6.0%) | 4,055 (19.9%) | 347 (23.1%)* | 5,838 (10.7%) | 116 (9.8%) |
| **Comorbidities, n (%)** | | | | | | |
| CHF | 6,182 (24.0%) | 304 (31.0%)* | 1,616 (7.9%) | 139 (9.3%) | 1,931 (3.5%) | 42 (3.5%) |
| COPD | 7,157 (27.8%) | 280 (28.5%) | 1,476 (7.2%) | 139 (9.3%)* | 1,952 (3.6%) | 36 (3.0%) |
| CVA | 3,625 (14.1%) | 155 (15.8%) | 613 (3.0%) | 83 (5.5%)* | 2,116 (3.9%) | 61 (5.1%)* |
| Malignancy | 1,607 (6.2%) | 55 (5.6%) | 407 (2.0%) | 23 (1.5%) | 168 (0.3%) | 2 (0.2%) |
| HIV | 138 (0.5%) | 5 (0.5%) | 39 (0.2%) | 5 (0.3%) | 7 (0.0%) | 2 (0.2%)* |
| Renal disease | 4,698 (18.2%) | 244 (24.8%)* | 1,116 (5.5%) | 93 (6.2%) | 1,295 (2.4%) | 23 (1.9%) |
| Liver disease | 2,034 (7.9%) | 131 (13.3%)* | 625 (3.1%) | 65 (4.3%)* | 218 (0.4%) | 6 (0.5%) |
| **Outcomes, n (%)** | | | | | | |
| Coma | 1,182 (4.6%) | 253 (25.8%)* | 873 (4.3%) | 514 (34.2%)* | 3,070 (5.6%) | 238 (20.0%)* |
| Mortality | 581 (2.3%) | 125 (12.7%)* | 837 (4.1%) | 286 (19.0%)* | 1,235 (2.3%) | 75 (6.3%)* |

Abbreviations: BMI: Body Mass Index; CCI: Charlson Comorbidity Index; CHF: Congestive Heart Failure; COPD: Chronic Obstructive Pulmonary Disease; CVA: Cerebrovascular Accident; HIV: Human Immunodeficiency Virus; IQR: interquartile range. *P-value < 0.05. P-values for continuous variables are based on pairwise Wilcoxon rank sum test. P-values for categorical variables are based on pairwise Pearson's chi-squared test for proportions.

## 3.2 Model Performance

The performance of DeLLiriuM in terms of AUROC compared to all baseline models is shown in Table II. The best structured EHR baseline model was the Transformer, with AUROC values of 0.84 (95% CI 0.81-0.87), 0.72 (95% CI 0.71-0.73), and 0.79 (95% CI 0.77-0.80) in internal, external 1, and external 2 validation sets. The best text EHR baseline was GatorTron-8b with AUROC values of 0.85 (95% CI 0.82-0.88), 0.74 (95% CI 0.73-0.75), and 0.80 (95% CI 0.79-0.81), respectively. The DeLLiriuM model showed higher performance than both baselines in the external validation sets, with AUROC values of 0.77 (95% CI 0.76-0.78) and 0.84 (95% CI 0.83-0.85) in the external validation sets 1 and 2. The receiving operator characteristic (ROC) curves for the best baselines and DeLLiriuM, as well as bar plots for AUROC for first seven days of admission are shown in Fig. 4. As seen in the figure, DeLLiriuM had higher performance than baseline models in both external sets and in most days.

**Table 2. DeLLiriuM performance compared to baseline models.**

| Modality | Model | Internal Validation Cohort (UF) | External Validation 1 Cohort (MIMIC) | External Validation 2 Cohort (eICU) |
|---|---|---|---|---|
| Structured EHR | NN | 0.84 (0.81-0.87) | 0.71 (0.70-0.72) | 0.68 (0.66-0.69) |
| | Transformer | 0.84 (0.81-0.87) | 0.72 (0.71-0.73) | 0.79 (0.77-0.80) |
| | Mamba | 0.83 (0.79-0.86) | 0.72 (0.71-0.73) | 0.72 (0.71-0.74) |
| Text EHR | ClinicalBERT | 0.82 (0.77-0.85) | 0.72 (0.71-0.74) | 0.78 (0.77-0.80) |
| | GatorTronS | 0.83 (0.80-0.86) | 0.74 (0.72-0.75) | 0.81 (0.80-0.82) |
| | GatorTron-8b | 0.85 (0.82-0.88) | 0.74 (0.73-0.75) | 0.80 (0.79-0.81) |
| | Llama 3-8b | 0.84 (0.81-0.87) | 0.74 (0.73-0.75) | 0.79 (0.77-0.80) |
| | DeLLiriuM | 0.85 (0.81-0.88) | **0.77 (0.76-0.78) [a,b]** | **0.84 (0.83-0.85) [a,b]** |

Abbreviations: AUROC: Area Under the Receiving Operating Characteristic; EHR: Electronic Health Records; NN: Neural Network. Performance is shown as the median AUROC across 200-iteration bootstrap with 95% Confidence Intervals in parenthesis. P-values are based on pairwise Wilcoxon rank sum tests. [a] p-value < 0.05 compared to best structured EHR baseline. [b] p-value < 0.05 compared to best text EHR baseline.

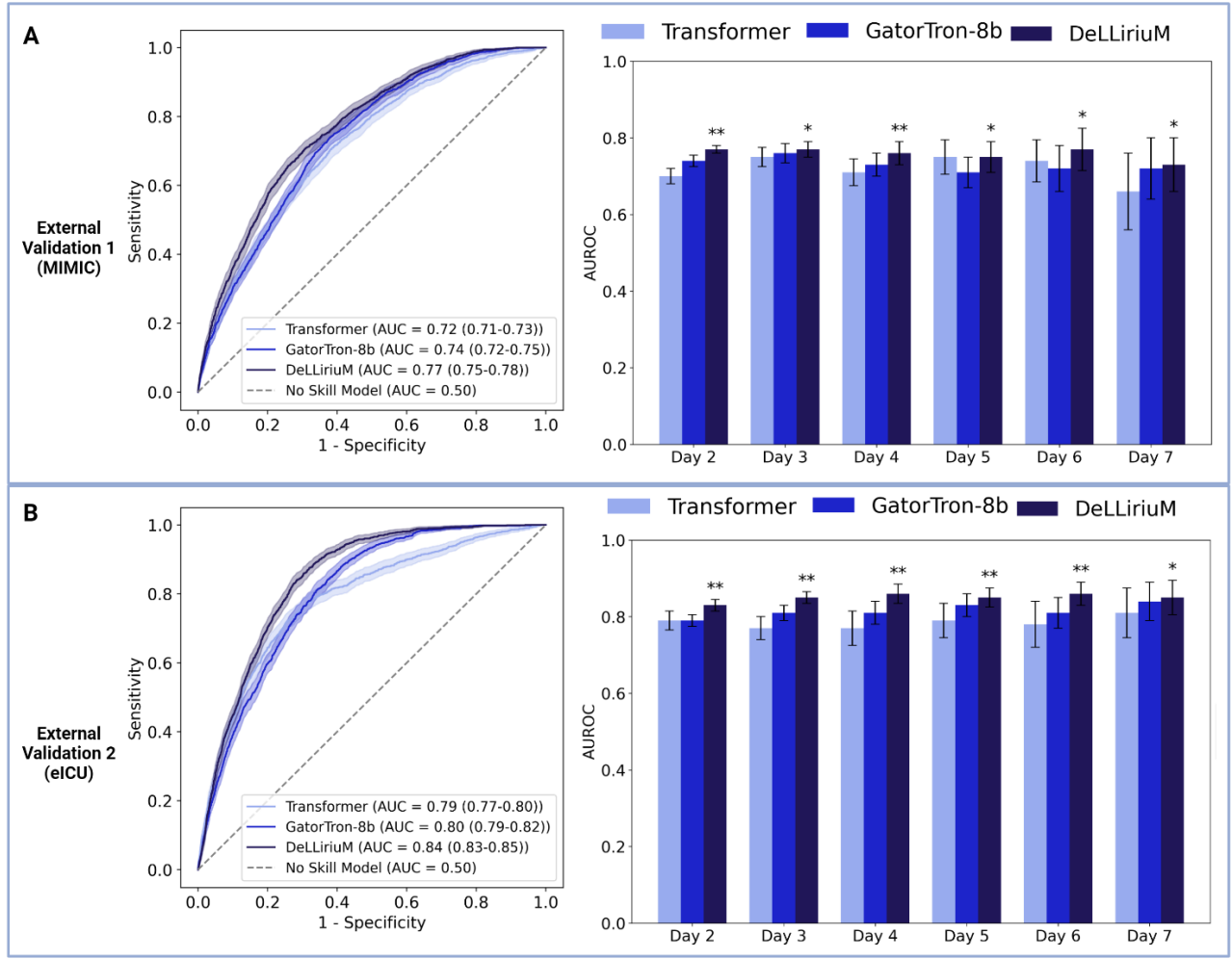

**Fig. 4 | DeLLiriuM performance.** ROC curves for overall performance and AUROC bar plots for first seven days of admission of best baseline models and DeLLiriuM on: (A) External Validation 1, (B) External Validation 2. *P-value < 0.05 compared to one baseline. **P-value < 0.05 compared to both. baselines.

3.3 Feature Importance

The top 15 features for prediction of delirium on each validation cohort according to absolute mean SHAP value, as well as top features in three subcategories (*i.e.,* vital signs, laboratory results, and comorbidities) are shown in Fig. 5. Among common features in the three cohorts, laboratory tests such as specific gravity of urine, brain natriuretic peptide (BNP), and anion gap were consistently within the top features. Ventilator settings such as total positive end expiratory pressure (PEEP) level and observed tidal volume also ranked highly. Vital signs such as heart rate, blood pressure, and oxygen saturation also had high SHAP values. Furthermore, SHAP text plots for four random samples drawn from the eICU cohort are shown in Fig. 6. Gradients of blue depict factors negatively impacting delirium, whereas gradients of red depict factors affecting delirium positively. The DeLLiriuM model recognized factors such as elevated creatinine and lactic acid levels, advanced age, abnormalities in vital signs, and more negative RASS scores (*i.e.,* higher levels of sedation) as predictive of delirium as shown by the positive impact on model output. On the other hand, high Glasgow coma scale (GCS) values, normal vital signs and laboratory values, and medication absence had a negative impact on model output.

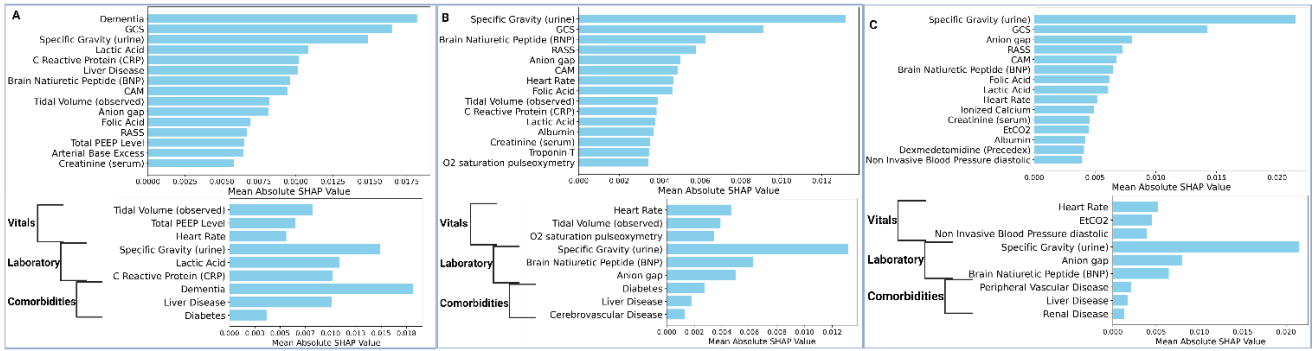

**Fig. 5 | SHAP analysis DeLLiriuM.** SHAP analysis bar plots for DeLLiriuM model in three validation cohorts: (A) Internal Validation (UFH), (B) External Validation 1 (MIMIC), (C) External Validation 2 (eICU).

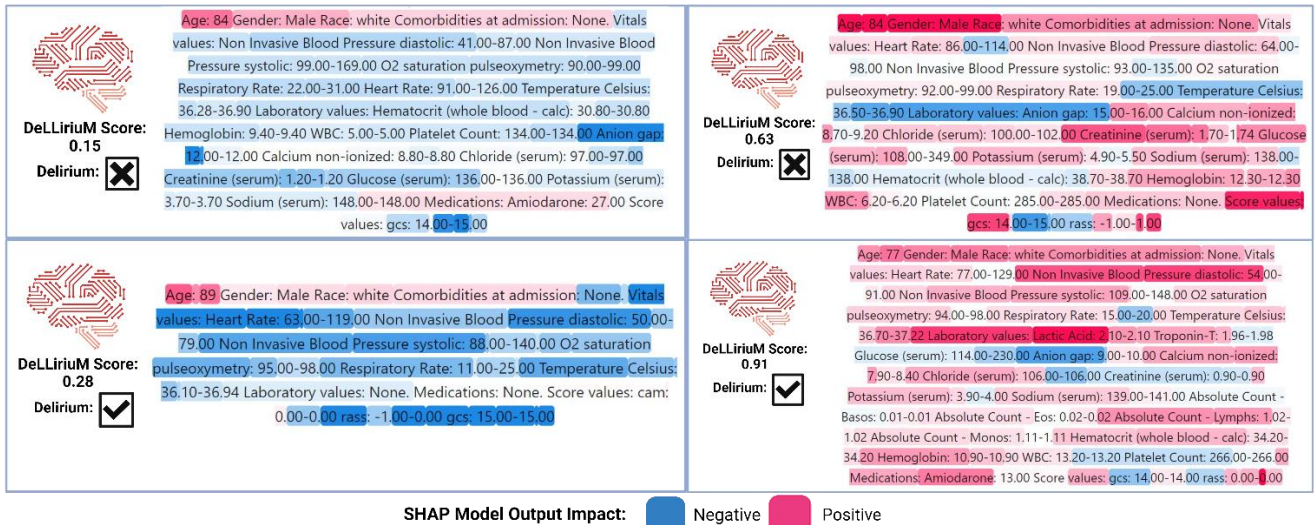

**Fig. 6 | DeLLiriuM predictions examples.** SHAP analysis text plots for DeLLiriuM model in random positive and negative delirium examples with low and high DeLLiriuM scores.

## 4. Discussion

4.1 Main Findings and Interpretation

The results presented in this study show the development and validation of DeLLiriuM as a tool for ICU patient delirium prediction. To the best of our knowledge, this is the first study to use structured EHR data in text form and LLMs for delirium prediction in the ICU. Previous studies to predict delirium using EHR data from the first 24 hours of ICU admission have shown varied performance. The PRE-DELIRIC model [6], developed on 1,613 patients from one hospital, achieved an AUROC of 0.84 (95% CI 0.82-0.87) in an external validation on 894 patients across four hospitals in the Netherlands. The E-PRE-DELIRIC model [7], developed on 1,962 patients from 13 ICUs across 7 countries, achieved an AUROC of 0.76 (95% CI 0.73-0.78). On the other hand, one study which used eICU data for developing a delirium prediction model, and MIMIC for external validation achieved an AUROC of 0.78 (95% CI 0.77-0.80) and 0.81 (95% C.I. not reported) on both datasets, respectively. Furthermore, a systematic review of ICU delirium prediction models revealed that AUROC varied greatly (from 0.62 to 0.94) across 23 different prediction models [36]. These studies have employed a varying number of features (anywhere from tens to hundreds of features) and a wide range of models (from logistic regression to deep learning models).

Although drawing direct comparisons between the performance of DeLLiriuM and these models is difficult due to the variation in development and validation settings, the results reported in this paper are comparable and even outperform what is seen in the literature. This study also develops and validates the proposed model with over 100,000 patients, the largest cohort compared to these studies and achieves high performance in both external validation sets spanning 194 hospitals while being developed on data from a single hospital. Furthermore, the DeLLiriuM model employs SOTA AI techniques and architectures with the use of an LLM and MLM domain-specific pre-training.

A few studies have employed LLMs with structured EHR data in text form for these tasks and have all seen improved performance compared to structured features deep learning approaches [20], [23]. These results are consistent with the results obtained in this work, where DeLLiriuM outperformed structured features (*i.e.,* structured EHR) deep learning approaches. This higher performance could be attributed to the large number of parameters that LLMs possess as well as the large corpus of text that they have been exposed to in their pre-training phase [20], [23].

A SHAP analysis for the three validation cohorts was conducted to understand how features in these datasets affect delirium prediction. The feature importance results are consistent with known delirium risk factors described in the literature. Acute hypoxemic respiratory failure (low blood oxygen levels) and acute hypercapnic respiratory failure (high blood carbon dioxide levels), especially in mechanically ventilated patients, are significant risk factors associated with delirium development in ICU patients [3], [37]. As a result, the physiologic parameters depicting signs of respiratory failure—low tidal volumes, higher PEEP levels, lower oxygen pulse oximetry levels and high end-tidal CO2 (EtCO2)—were significant features to predict delirium across the three cohorts. Similarly, sepsis and septic shock due to pneumonia, urinary tract infections, bloodstream infection are well-known etiologic factors in ICU delirium development. Septic patients typically present with tachycardia (high heart rates), have elevated lactic acid and C-reactive protein (CRP) levels and sometimes elevated anion gaps, thereby explaining the importance of these features in the three cohorts [3], [37]. Patients with systemic conditions like dementia, diabetes, liver disease and renal disease are highly predisposed to develop delirium [3], [37].

Although not typically known as a risk factor for delirium, abnormal urine specific gravity values emerged as a consistent top laboratory feature of importance across the three cohorts for delirium prediction. The relevance of this feature could be attributed to the fact that patients with abnormal urine specific gravity may present with a variety of metabolic abnormalities (abnormal sodium, calcium, magnesium, glucose levels, dehydration) [38] as well as urinary tract infection [39], leading to confusion, mental status changes and delirium development. Lin et al. [40] found that urine specific gravity greater than 1.010 predicted early neurologic deterioration in patients with acute ischemic stroke. Another study by Kim et al. [41] concluded that urine specific gravity above 1.030 was a statistically significant factor for delirium development in post-operative general surgery patients.

Four random examples drawn from the eICU cohort provided demonstrations of the model's ability to recognize delirium risk factors (Fig. 6). The DeLLiriuM model was able to identify positive predictors of delirium such as advanced age, elevated laboratory values such as creatinine and lactic acid, and abnormalities in vital signs such as low diastolic and systolic blood pressure. On the other hand, it recognized negative predictors of delirium development such as normal laboratory results, normal physiologic values (GCS, oxygen saturation) and vital signs (heart rate, blood pressure). Two of the examples show cases where DeLLiriuM predicted a high score (0.63) for a patient that did not develop delirium and a low score (0.28) for a patient that developed delirium. In the first case, the patient had clear risk factors for delirium, such as advanced age, elevated creatinine levels, and low hemoglobin which can explain the high predicted score. In the second case, the patient had no reported laboratory values and normal values for some vital signs which can also explain the low predicted score. These findings are all consistent with the literature [37]. Although some inconsistencies are seen in the analysis (*e.g.,* GCS score of 14 having positive impact on one example and negative impact on another), it is important to consider the interactions between features that could lead to different impacts on delirium risk.

4.2 Limitations and Future Work

This study is not without limitations. First, the short context length of GatorTronS, 512, limits the amount of information that can be included in the EHR text report. Out of the tested models, LLaMa 3 had a longer context length (2,048). However, the overall performance of this model did not exceed that of DeLLiriuM. This could be explained due to GatorTronS being a clinical LLM, trained on a large corpus of clinical text, whereas LLaMa 3 is a general LLM. Future experiments will include use of other LLMs with longer context lengths as well as use of approaches such as FlashAttention [42].

The summarization method also has its limitations. Using only the minimum and maximum value of each temporal feature might exclude relevant temporal dynamics of these variables. Future experiments will explore other summarization approaches such as multimodal models which could use text and structured features [43], and using LLMs such as LLaMa3-70b and GPT-4 [44] for more comprehensive summaries.

Finally, the development and validation of DeLLiriuM was retrospective. Therefore, future work will focus on validation of the model on prospective cohorts to measure its performance on real-world settings. Furthermore, the DeLLiriuM model will be extended to make continuous risk prediction of delirium to provide a real-time monitoring of patients' mental status.

# 5. Conclusion

In summary, we have developed and validated retrospectively DeLLiriuM, an LLM-based model for prediction of delirium in the ICU after the first 24 hours of admission. Our DeLLiriuM model demonstrates superior performance compared to deep learning models that utilize comprehensive sequential EHR data, despite the fact that it only uses a summary of the sequential data as input. This further underscores the proficiency of LLMs in capturing the nuances and subtleties present in the data.

DeLLiriuM allows for delirium risk screening of patients and can provide helpful information to clinicians to make timely interventions. Furthermore, we proposed an automatic text report generation method for structured EHR data, a novel approach for text classification interpretability, and training procedures for clinical outcome predictions using LLMs that can be applied to other prediction tasks. The DeLLiriuM model will be validated prospectively to measure its performance in real-world settings and extended to continuous risk prediction to provide a real-time monitoring of patients' mental status.

# 6. Acknowledgement

A.B, P.R., and T.O.B. were supported by NIH/NINDS R01 NS120924, NIH/NIBIB R01 EB029699. PR was also supported by NSF CAREER 1750192.